\documentclass[lettersize,journal]{IEEEtran}
\usepackage{amsmath,amsfonts,amssymb}
\usepackage{algorithm}
\usepackage{array}
\usepackage[caption=false,font=normalsize,labelfont=sf,textfont=sf]{subfig}
\usepackage{textcomp}
\usepackage{stfloats}
\usepackage{url}
\usepackage{verbatim}
\usepackage{graphicx}
\usepackage{cite}
\newcommand{\figref}{Fig.~\ref}
\usepackage[skins]{tcolorbox}

\usepackage{algpseudocode}
\usepackage{xcolor}
\usepackage{hyperref}
\usepackage{multirow}
\usepackage{booktabs, caption}
\usepackage{tabularx}
\usepackage{float}

\usepackage{colortbl}
\usepackage{graphicx}

\usepackage{xcolor}

\newcommand\rebuttalthird [1]{\textcolor{black}{#1}}

\newcommand\rebuttal [1]{\textcolor{black}{#1}}

\newcommand\rebuttalsecond [1]{\textcolor{black}{#1}}
\newtcolorbox{takeawaybox}[2][]{
  enhanced,
  attach boxed title to top left={yshift=-0.1in, xshift=0.15in},
  colback=phaseOneBlue!10,
  colframe=phaseOneBlue,
  colbacktitle=phaseOneBlue!80!black,
  fonttitle=\bfseries,
  title=#2,
  #1
}

\definecolor{phaseOneBlue}{RGB}{70, 100, 170}
\definecolor{phaseTwoRed}{RGB}{192, 28, 40}

\renewcommand{\arraystretch}{1.1}
\newcommand{\best}[1]{\cellcolor{gray!20}\textbf{#1}}

\begin{document}
\title{Pruning  Federated Models through Loss Landscape Analysis and Client Agreement Scoring}


\author{
    Christian Internò, 
    Elena Raponi,
    Markus Olhofer, 
    Ali Raza,\\
    Thomas Bäck, 
    Niki van Stein, 
    Yaochu Jin, 
    and Barbara Hammer
    \thanks{This work was co-advised by B. Hammer and E. Raponi.}
    \thanks{Corresponding author: Christian Internò (c.interno@uni-bielefeld.de).}
    \thanks{C. Internò and B. Hammer are with Bielefeld University, Germany; C. Internò, M. Olhofer, and A. Raza are with the Honda Research Institute Europe, Germany; Y. Jin is with the School of Engineering, Westlake University, China; E. Raponi, T. Bäck, and N. van Stein are with the Leiden Institute of Advanced Computer Science (LIACS), Leiden University, Netherlands.}
\thanks{This is the accepted version of the article published in the IEEE Internet of Things Journal, 2026, doi: \href{https://doi.org/10.1109/JIOT.2026.3686028}{10.1109/JIOT.2026.3686028}. \textcopyright~2026 IEEE. }
}

\maketitle

\begin{abstract}
The practical deployment of Federated Learning (FL) on resource-constrained devices is fundamentally limited by the high cost of training large models and the instability caused by heterogeneous (non-IID) client data. Conventional pruning methods often treat data heterogeneity as a problem to be mitigated. In this work, we introduce a paradigm shift: we reframe client diversity as a feature to be harnessed. We propose \texttt{AutoFLIP}, a framework that begins not with training, but with a one-time \textit{federated loss exploration}. During this phase, clients collaboratively build a map of the collective loss landscape, using their diverse data to reveal the problem's essential structure. This shared intelligence then guides an adaptive pruning strategy that is dynamically refined by \textit{client agreement} throughout training. This approach allows \texttt{AutoFLIP} to identify robust and efficient sub-networks from the outset. Our extensive experiments show that \texttt{AutoFLIP} reduces computational overhead by an average of 52\% and communication costs by over 65\% while simultaneously achieving state-of-the-art accuracy in challenging non-IID settings.
\end{abstract}

\begin{IEEEkeywords}
Federated Learning, Model Compression, Loss Exploration, Non-IID Data, Deep Learning, Edge AI. 
\end{IEEEkeywords}

\section{Introduction}
\IEEEPARstart{T}{he} proliferation of intelligent edge devices presents a fundamental conflict between the increasing complexity of deep neural networks (DNNs) and the limited resources of the hardware on which they must operate \cite{8808168, 8869705}. Federated Learning (FL) has emerged as a leading paradigm for training models in such distributed, privacy-sensitive environments \cite{zhu2021federated, Rieke2020}. However, its practical application is severely constrained by two interconnected challenges: the high communication and computational costs of large models, and the statistical heterogeneity (non-IID data) across clients, which can destabilize training and degrade model performance \cite{kairouz2021advancesopenproblemsfederated, 10.1145/3581759}.

Here, we introduce \texttt{AutoFLIP}, which offers a solution to this conflict by fundamentally rethinking how model pruning is performed in federated settings. Our approach decouples the generation of pruning criteria from the iterative process of weight training. We introduce an explicit, one-time \textit{federated exploration phase} designed not to train the model, but to generate metadata about the collective loss landscape. During this phase, clients probe their local data to create maps of parameter loss sensitivity, which the server aggregates into a global map of parameter importance. This collaboratively-built intelligence then guides an adaptive hybrid pruning in all subsequent training rounds.
\begin{figure}[!t]
    \centering
    \includegraphics[width=1\columnwidth]{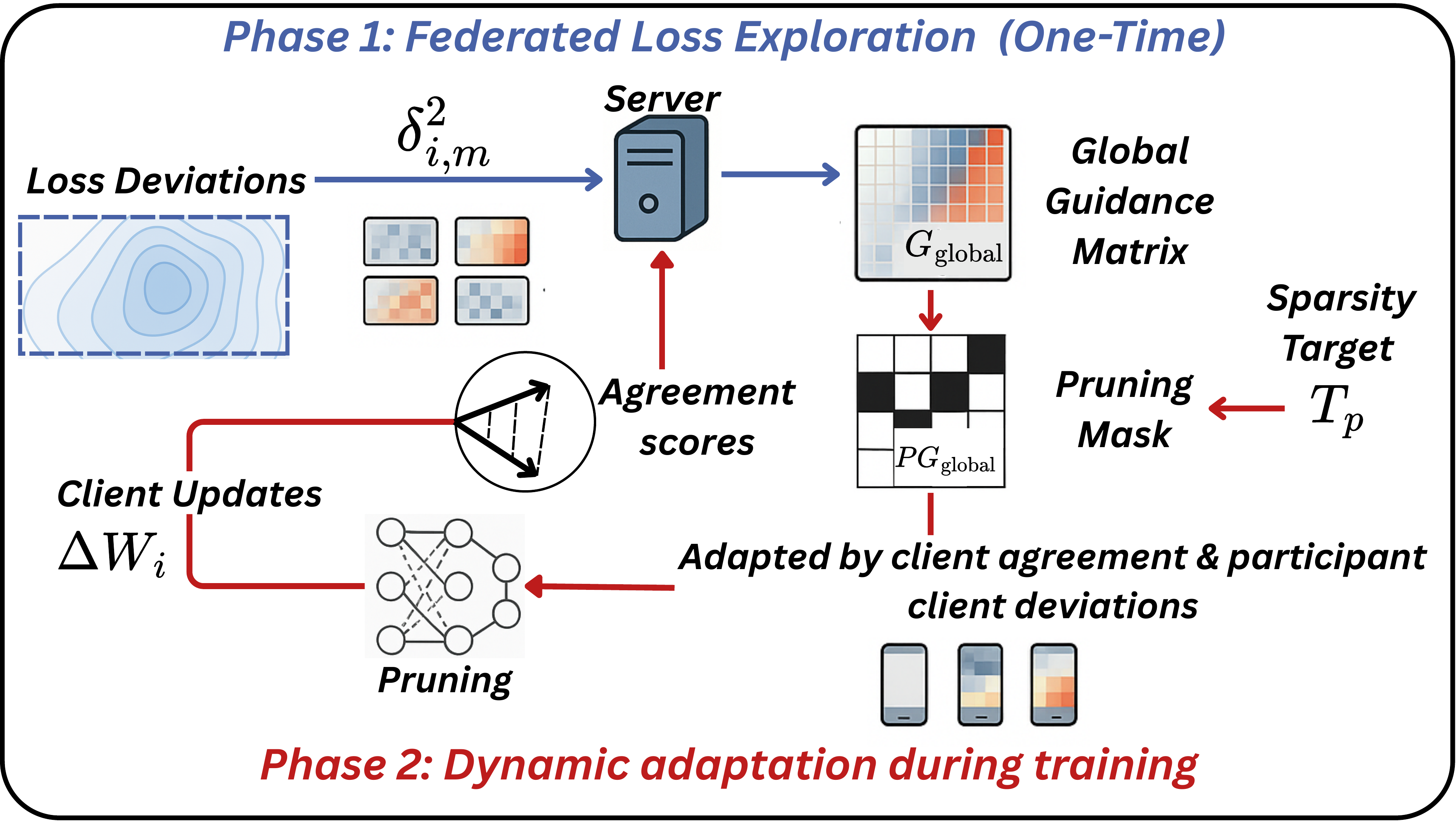}
    \caption{Conceptual overview of the \texttt{AutoFLIP} framework. \textcolor{phaseOneBlue}{\textbf{Phase 1 (One-Time):}} An initial \textit{Federated Loss Exploration}, where clients provide loss landscape deviations to construct the Global Guidance Matrix. \textcolor{phaseTwoRed}{\textbf{Phase 2 (Continuous):}} A \textit{Dynamic Adaptation} phase, where the guidance matrix is continuously refined using participant client agreement scores. The evolved matrix is then used to generate an adaptive pruning mask.}

    \label{fig:autoflip_workflow}
\end{figure}

Crucially, our framework leverages the ongoing stream of client updates to dynamically refine this guidance, turning the challenge of data heterogeneity into a source of intelligence. By identifying parameters that are not only salient but also subject to high client agreement, \texttt{AutoFLIP} finds a robust, common sparse sub-network. This unified approach, illustrated in Figure~\ref{fig:autoflip_workflow}, allows for both fine-grained unstructured pruning to reduce communication costs and emergent structured pruning of entire components (e.g., neurons, filters) to physically reduce on-device FLOPs. The process is governed by a single pruning threshold, $T_p$, which allows the system to translate a high-level intent for efficiency (e.g., ``reduce model size by 40\%'') into an optimized hardware-aware configuration, a key concept in Intent-Based Networking (IBN) \cite{IBN_Survey}.

\rebuttal{Specifically, the pruning threshold $T_p$ serves as a direct \textit{intent interface}, allowing an operator to set a high-level declarative policy (e.g., ``reduce resource usage by 40\%''). \texttt{AutoFLIP} then acts as the autonomous control loop, translating this intent into an optimized, hardware-aware sub-network configuration at the edge without requiring manual, low-level intervention. This aligns directly with the core IBN principles of automated configuration and intent-driven operation~\cite{11004066}.}

\rebuttal{This two-pronged approach—a one-time loss exploration followed by agreement-based refinement—fundamentally differentiates \texttt{AutoFLIP} from existing SoTA methods. Unlike strategies that rely on standard scores or learn divergent, personalized masks, our framework is uniquely designed to collaboratively discover a \textit{single} robust and efficient sub-network from the collective data loss. A detailed discussion with SoTA works such as \texttt{PruneFL} and \texttt{FedMask} is presented in Section~\ref{sec:background}.} Our primary contributions are:
\begin{enumerate}
    \item A novel framework, \texttt{AutoFLIP}, that formally connects FL with IBN by translating a high-level intent for model efficiency into a concrete, optimized, and pruned model configuration at the network edge.
    \item A dynamic pruning mechanism featuring: (a) an enhanced guidance metric that incorporates both parameter saliency and client update agreement, and (b) a temporally-aware pruning, where the mask evolves throughout training to adapt to the loss changing.
    \item A comprehensive empirical study showing that \texttt{AutoFLIP} reduces computational overhead (by an average of 52\%) and accelerates convergence (reducing communication costs by over 65\% in comparable scenarios) while simultaneously improving final model accuracy on challenging non-IID benchmarks.
\end{enumerate}
Code for implementation: \url{https://github.com/ChristianInterno/AutoFLIP}.\\

\noindent\textbf{Structure of the Paper:} 
 Section~\ref{sec:background} provides related work on model pruning in both centralized and federated settings. Section~\ref{sec:preliminaries} establishes the key notations and problem statement. We detail the proposed \texttt{AutoFLIP} methodology in Section~\ref{sec:methodology}. Section~\ref{subsec:exp} presents the experimental results, followed by \rebuttal{robustness to noisy clients resistance in Section~\ref{sec:noyse}} and ablation studies in Section~\ref{sec:ablation}. Finally, we conclude the paper and outline future research directions in Section~\ref{sec:conclusion}.

\section{Background and Related Work}
\label{sec:background}

\subsection{Pruning in Deep Learning}
Model pruning is a key technique for reducing the computational requirements of Deep Learning (DL) models by removing non-essential parameters \cite{NIPS198807e1cd7d, NIPS19896c9882bb}. Most approaches prune a pre-trained network by scoring parameters based on metrics like weight magnitude \cite{Han2015ccm} or Taylor expansions \cite{DBLP:conf/iclr/MolchanovTKAK17, NIPS19896c9882bb}. While effective for inference, these centralized methods are ill-suited for federated applications due to data privacy constraints and poor adaptation to diverse local data \cite{you_gate_2019,lin_hrank_2020}. Other directions like pruning at initialization \cite{lee2018snip} are data-agnostic, while dynamic sparse training \cite{pmlr-v119-evci20a} is often too memory-intensive for resource-constrained edge devices.

\subsection{Pruning in Federated Learning}
From an IBN perspective, the ability to automatically prune models is a core mechanism for on-device efficiency. However, applying pruning techniques in FL environments presents unique challenges. While the standard \texttt{FedAvg} algorithm \cite{mcmahan2023communicationefficient} offers no model compression, recent works have begun to integrate pruning into the federated workflow.

The current SoTA includes methods like \texttt{PruneFL} \cite{jiang_model_2023}, which focus on unstructured (weight-level) pruning, and \texttt{EFLPrune} \cite{Wu2023}, which employs a hybrid strategy. Other recent works have also explored learning a single global mask. For instance, \texttt{FedMask} \cite{fedmask} aims for personalization by learning a unique, heterogeneous mask for each client. While this can tailor a model to a specific client's data, it does so by creating divergent sub-networks, which can hinder the convergence towards a single, powerful global model.

\rebuttal{Although \texttt{AutoFLIP} is the first to formally connect FL with neural pruning and IBN, the SoTA baselines can also be framed as IBN-inspired, as they fulfill an implicit ``intent'' through automated pruning. Methods like \texttt{PruneFL}~\cite{jiang_model_2023} fulfill an \textit{intent for efficiency}, as they have an implicit intent of ``minimizing training time.'' They \textit{translate} this intent using heuristic-based scores (e.g., weight magnitude), resulting in unstructured pruning that saves communication but not necessarily on-device computation (FLOPs). On the other hand, methods like \texttt{FedMask}~\cite{fedmask} have an \textit{intent for personalization} on a ``client-level.'' They \textit{translate} this by autonomously learning unique, heterogeneous masks for each client.}

Our method, \texttt{AutoFLIP}, diverges from these approaches in two critical ways. First, in contrast to methods that are purely unstructured, like \texttt{PruneFL}, our unified hybrid mechanism can physically reduce FLOPs through emergent structured pruning. Second, and more fundamentally, our pruning decisions are not guided by standard scores or joint optimization. Instead, they are informed by our novel \textit{federated loss exploration phase}. This makes our approach uniquely data-driven and collaborative, using the collective client experience as the primary signal for optimization. We identify \texttt{PruneFL}, \texttt{EFLPrune}, and the recent \texttt{FedMask} as the key SOTA baselines for our experimental comparison.

\section{Preliminaries}
\label{sec:preliminaries}

\begin{table}[h]
\caption{Summary of Notations}
\label{tab:notations}
\centering
\small
\renewcommand{\arraystretch}{0.9}
\setlength{\tabcolsep}{4pt}
\begin{tabular}{ll}
\toprule
\textbf{Symbol} & \textbf{Description} \\
\midrule
\( C, K, R, E \) & Total clients, participants per round, rounds, epochs \\
\( W_{\text{global}}, W_i \) & Global and local model parameters \\
\( \Delta W_i \) & Parameter update from client \( i \) \\
\( \sigma^2_{\Delta W} \) & Variance of client update vectors \\
\midrule
\( T_p \) & Pruning threshold \\
\( G_{\text{global}} \) & Global Guidance Matrix \\
\( \delta_{i,m}^2 \) & Initial squared parameter deviation \\
\( V_m^{(r)} \) & Variance of updates for parameter $m$ in round $r$ \\
\( A_m^{(r)} \) & Client Agreement Score \\
\( I_m^{(r)} \) & Round-specific Importance Score \\
\( PG_{\text{global}}^{(r)} \) & Binary pruning mask for round $r$ \\
\bottomrule
\end{tabular}
\end{table}

\subsection{Federated Learning}
\label{sec:fl_preliminaries}
Table~\ref{tab:notations} defines the primary symbols used in this paper.

\begin{figure}[h]
    \centering
    \includegraphics[width=\columnwidth]{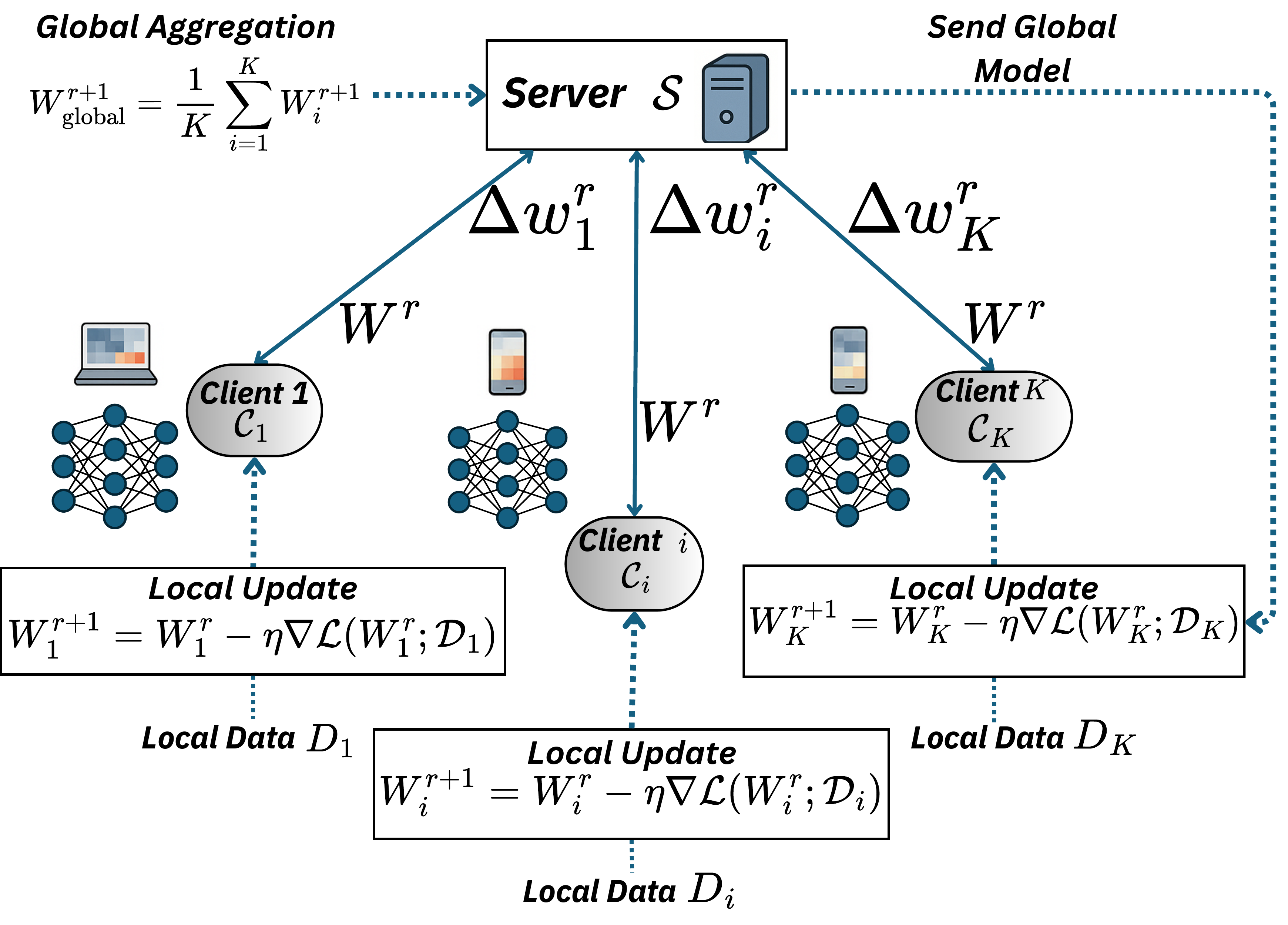}
    \caption{FL optimization process. At each communication
round, participant clients perform local updates to then send the new parameters to the Server for the global aggregation}
    \label{fig:opt}
\end{figure}

FL is a distributed machine learning paradigm where multiple clients collaboratively train a shared global model, $W_{\text{global}}$, without centralizing their private data. Each client $i$ possesses a local dataset $D_i$ drawn from a local data distribution $p_i$, as detailed in Figure \ref{fig:opt}.

In real-world scenarios, these distributions are typically non-independent and identically distributed (non-IID). The global objective is to minimize a loss function, $\mathcal{L}$, averaged over all clients:
\begin{equation}
    \min_{W_{\text{global}}} \mathcal{L}(W_{\text{global}}) := \frac{1}{C} \sum_{i=1}^{C} \mathcal{L}_i(W_{\text{global}}),
\end{equation}
where $\mathcal{L}_i(W_{\text{global}}) = \mathbb{E}_{(x, y) \sim p_i}[\mathcal{L}(W_{\text{global}}, x, y)]$ is the local objective for client $i$.

The standard FL process, often realized through the Federated Averaging (\texttt{FedAvg}) algorithm, is iterative. In each communication round $r$, (1) the server selects a subset of $K$ clients and distributes the current global model $W_{\text{global}}^{(r-1)}$; (2) each client performs local training for $E$ epochs to produce an updated local model $W_i^{(r)}$; and (3) the server aggregates the client updates, typically by averaging, to form the new global model $W_{\text{global}}^{(r)}$.

\subsection{Key Challenges and Objectives}
\label{sec:objectives}

The effectiveness of FL is hindered by three primary challenges. First, the statistical heterogeneity of client data (non-IID) causes significant variance among client updates ($\sigma^2_{\Delta W}$), which can destabilize and slow the convergence of the global model. Second, the need to frequently transmit large model updates between clients and the server imposes a high communication cost ($C_{\text{comm}}$). Third, the computational cost ($C_{\text{comp}}$) of training modern deep learning models can be prohibitive for resource-constrained edge devices.

The objective of \texttt{AutoFLIP} is to address these three challenges simultaneously. We aim to design a pruning strategy, $\mathcal{P}$, that dynamically removes non-essential parameters to jointly minimize client update variance, communication overhead, and on-device computational load:
\begin{equation}
    \text{Find } \mathcal{P} \text{ to minimize } (\sigma^2_{\Delta W}, C_{\text{comm}}, C_{\text{comp}}).
\end{equation}
Our approach achieves this by generating a data-driven pruning mask that adapts to the state of the model and the federated network throughout training.
\section{Methodology}
\label{sec:methodology}

The \texttt{AutoFLIP} framework prunes neural networks during federated training by dynamically identifying and removing non-essential parameters. The core of our method is a guidance system that continuously learns from client update behavior to inform its pruning decisions. The framework operates in two primary phases: (1) an initial, one-time exploration phase to generate a foundational map of parameter loss, and (2) a continuous adaptation phase that refines this map during each training round by assessing client update agreement. This two-stage process ensures the pruning strategy remains effective throughout training.

\subsection{Adaptive Guidance Generation}
\label{subsec:adaptive_guidance}

\texttt{AutoFLIP} constructs and maintains a global guidance matrix, $G_{\text{global}}$, which quantifies the federated consensus on parameter saliency. This matrix is initialized once and then dynamically updated in each communication round.

The process begins with a subset of clients ($C_\text{exp}$) performing an initial federated loss exploration to create the foundational guidance map. Each client $i \in C_\text{exp}$ receives the initial global model, $W_\text{global}^{(0)}$, and trains it for $E_\text{exp}$ local epochs. Rather than the final weights, we use the squared deviation of parameters as a proxy for its sensitivity to local loss. For each parameter $m$, we measure the squared deviation from the initial state:
\begin{equation}
    \delta_{i,m}^2 = (W_{i,m}^{(E_\text{exp})} - W_{i,m}^{(0)})^2.
    \label{eq:deviation}
\end{equation}
The server aggregates these scores, computes the average deviation $\bar{\delta}_m^2$ for each parameter, and normalizes these values to form the initial global guidance matrix, $G_{\text{global}}^{(0)}$:
\begin{equation}
    G_{\text{global}, m}^{(0)} = \frac{\bar{\delta}_m^2 - \min_j(\bar{\delta}_j^2)}{\max_j(\bar{\delta}_j^2) - \min_j(\bar{\delta}_j^2)}.
    \label{eq:norm_deviation}
\end{equation}

\rebuttalthird{We justify the use of $\delta_{i,m}^{2}$ via the second-order Taylor expansion of the loss function, a standard method for estimating parameter saliency~\cite{NIPS19896c9882bb, molchanov2019importance}. We assume that after $E_{exp}$ epochs of exploration, the local client models approach a local minimum where the first-order gradient term vanishes ($g \approx 0$). Under this condition, the change in loss $\delta\mathcal{L}$ is dominated by the second-order term:}
\begin{equation}
\delta\mathcal{L} \approx \frac{1}{2} \Delta w^T H \Delta w \implies \delta\mathcal{L}_m \approx \frac{1}{2} h_m \underbrace{(\Delta w_m)^2}_{\equiv \delta_{i,m}^2}
\end{equation}
\rebuttalthird{where $H$ is the Hessian matrix and $h_m$ is the curvature for parameter $m$. The underbraced term highlights that our metric $\delta_{i,m}^2$ is the direct proxy for the quadratic displacement term. While explicit computation of $h_m$ is computationally infeasible for resource-constrained IoT devices, this relationship demonstrates that saliency is effectively captured by the squared displacement. Large deviations imply high sensitivity, while minimal displacements indicate ``flat'' regions safe for pruning~\cite{NEURIPS2019f34185c4}.}

\rebuttalthird{We note that while explicit computation of curvature $h_m$ (or Fisher Information~\cite{pmlrkarakida19a}) might be feasible in high-resource cross-silo settings with larger computation, it is computationally prohibitive for the \textit{cross-device} IoT scenarios targeted by AutoFLIP. Consequently, we rely on the squared deviation as the most efficient proxy to approximate structural sensitivity without incurring additional memory or compute overheads.}

\rebuttal{This \textit{Federated Loss Exploration}~\cite{10651455} phase functions as a form of meta-learning \cite{Thrun1998, pmlr-v70-finn17a} where the network learns an \textit{inductive bias} \cite{Mitchell2007TheNF}—the global guidance matrix—that enables more efficient downstream training. We use squared parameter deviation as a proxy for loss sensitivity \cite{neco} to collaboratively ``map'' the landscape, distinguishing flat, robust regions (low deviation, safe to prune) from sharp, sensitive regions (high deviation, critical to keep). This process is conceptually analogous to Guided Transfer Learning~\cite{nikolić2023guided, li2023usingguidedtransferlearning, 10651455}, where ``scouts'' transfer knowledge on diverse local sub-problems.}

A static guidance map can become obsolete as the model state evolves. \texttt{AutoFLIP} , therefore, refines $G_{\text{global}}$ in every training round by calculating an importance score that incorporates inter-client agreement.
A globally beneficial parameter update should exhibit strong consensus among clients. We define a holistic Client Agreement Score, $A_m^{(r)}$, that jointly assesses two components of consensus: directional alignment and magnitude consistency. For each parameter $m$ in round $r$, the score is:
\begin{equation}
    A_m^{(r)} =  \underbrace{\left| \frac{1}{K} \sum_{i \in \mathcal{K}_r} \text{sign}(\Delta W_{i,m}^{(r)}) \right|}_{\text{Directional Agreement}} \cdot \underbrace{\frac{1}{1 + V_m^{(r)}}}_{\text{Magnitude Agreement}},
    \label{eq:agreement}
\end{equation}
where $K$ clients are participating, $\Delta W_{i,m}^{(r)}$ is the parameter update from client $i$, and $V_m^{(r)}$ is the variance of those updates. The score $A_m^{(r)}$ approaches 1 only when clients agree on both the direction of an update and its magnitude.

At the end of each round $r$, the server computes a round-specific importance score, $I_m^{(r)}$. To do this, it first calculates the magnitude of the aggregated global update from the participating clients, $(\Delta W_m^{(r)})^2$. It then normalizes these values across all parameters using only the statistics from the current round to obtain the normalized magnitude $\hat{\Delta}_m^{(r)}$. This score is then combined with the agreement score, $A_m^{(r)}$, to amplify the importance of updates that have strong consensus:

\begin{equation}
    I_m^{(r)} = \hat{\Delta}_m^{(r)} \cdot (1 + A_m^{(r)}).
    \label{eq:importance}
\end{equation}
Instead of a simple cumulative update, we use an adaptive, self-tuning mechanism to integrate this new information. The Client Agreement Score $A_m^{(r)}$ serves as a dynamic, per-parameter learning rate in an Exponential Moving Average (EMA). This allows the system to place more trust in new information when consensus is high, and rely on historical guidance when consensus is low:
\begin{equation}
    G_{\text{global}, m}^{(r)} = (1 - A_m^{(r)}) G_{\text{global}, m}^{(r-1)} + A_m^{(r)} I_m^{(r)}.
    \label{eq:adaptive_ema_update}
\end{equation}
This data-driven trust mechanism ensures the guidance matrix evolves robustly, balancing stability with adaptability to the changing model state.

\begin{takeawaybox}{Takeaway 1: Novel Mechanism}
Our pruning is not random. Instead, it is guided by two signals: a collaborative map of the loss landscape, and the consensus among clients on how that parameter should change. We only keep parameters that are both loss important and in agreed consensus.
\end{takeawaybox}

\subsection{Intent-Driven Hybrid Pruning}
\label{subsec:hybrid_pruning}
The pruning mechanism uses the global guidance matrix, $G_{\text{global}}^{(r-1)}$, to generate a binary pruning mask, $PG_{\text{global}}^{(r)}$, for the current round. This process is governed by a pruning threshold, $T_p \in [0, 1]$, which enables direct control over model sparsity. The mask is generated by thresholding the guidance matrix:
\begin{equation}
    PG_{\text{global},m}^{(r)} = 
    \begin{cases} 
        1 & \text{if } G_{\text{global},m}^{(r-1)} \geq T_p \\
        0 & \text{otherwise}
    \end{cases}
    \label{eq:binarization}
\end{equation}
Applying this single mask achieves a powerful hybrid pruning effect, simultaneously inducing unstructured sparsity for communication efficiency and enabling structured pruning for computational savings.

\subsubsection{Unstructured Pruning for Communication Efficiency}
\begin{figure}[!t]
    \centering
    \includegraphics[width=0.95\columnwidth]{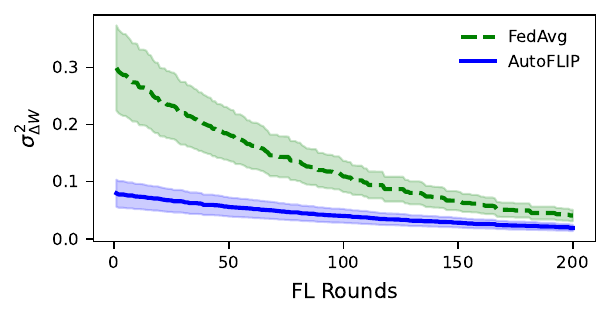}
    \caption{Variance of weight updates ($\sigma^2_{\Delta W}$) over FL rounds for \texttt{AutoFLIP} and \texttt{FedAvg}, illustrating the stabilizing effect of guided pruning.}
    \label{fig:variance_reduction}
\end{figure}

The immediate effect of the mask is fine-grained unstructured pruning, where individual parameters are set to zero if their guidance score is below $T_p$. This reduces the number of non-zero parameters that must be communicated, lowering communication overhead. By constraining updates to a common sparse subnetwork, this strategy also reduces the dimensionality of client disagreement, leading to a substantial decrease in the variance of weight updates and promoting stable convergence (Fig.~\ref{fig:variance_reduction}).

\subsubsection{Emergent Structured Pruning for Computational Efficiency}
\begin{figure*}[!t]
    \centering
    \includegraphics[width=\textwidth]{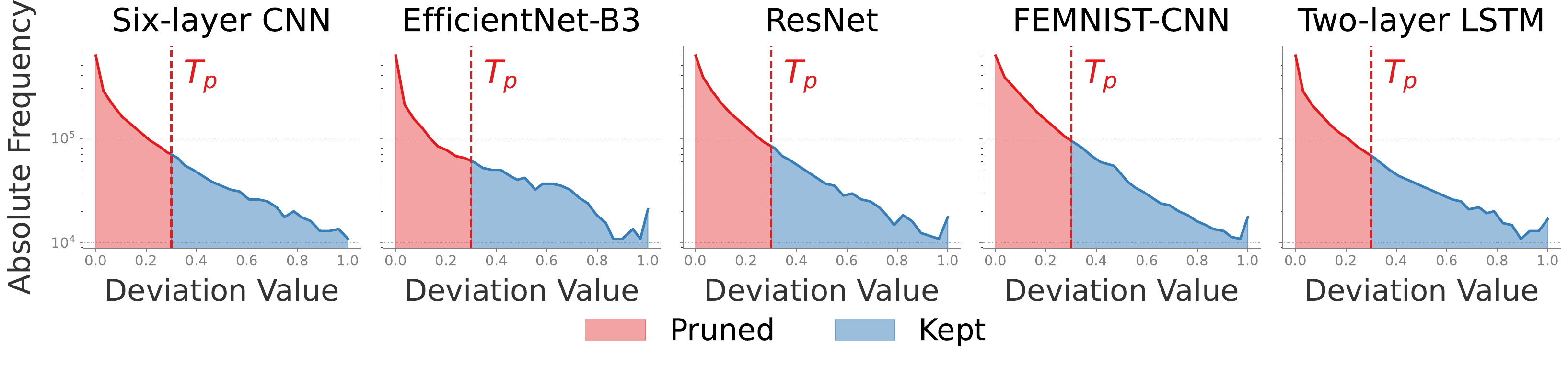}
    \caption{Distribution of $G_{\text{global}}$. The absolute frequency (in log-scale) is shown for
each normalized deviation. Higher frequencies are recorded for smaller deviation values, indicating that many parameters are
irrelevant for loss improvement.}
    \label{fig:G_global_distr}
\end{figure*}

While unstructured sparsity reduces data transfer, it does not inherently decrease the computational load (FLOPs) on standard hardware. \texttt{AutoFLIP} achieves this through \textit{emergent structured pruning}. When the mask removes all parameters associated with a specific structural unit $S$ (e.g., a neuron or a convolutional filter), the entire unit is deactivated. This removal of entire components from the network graph is the key to computational savings. The final distribution of guidance scores confirms that the framework effectively identifies a large subset of non-essential parameters for pruning (Fig.~\ref{fig:G_global_distr}).

\begin{figure}[!t]
    \centering
    \includegraphics[width=0.8\columnwidth]{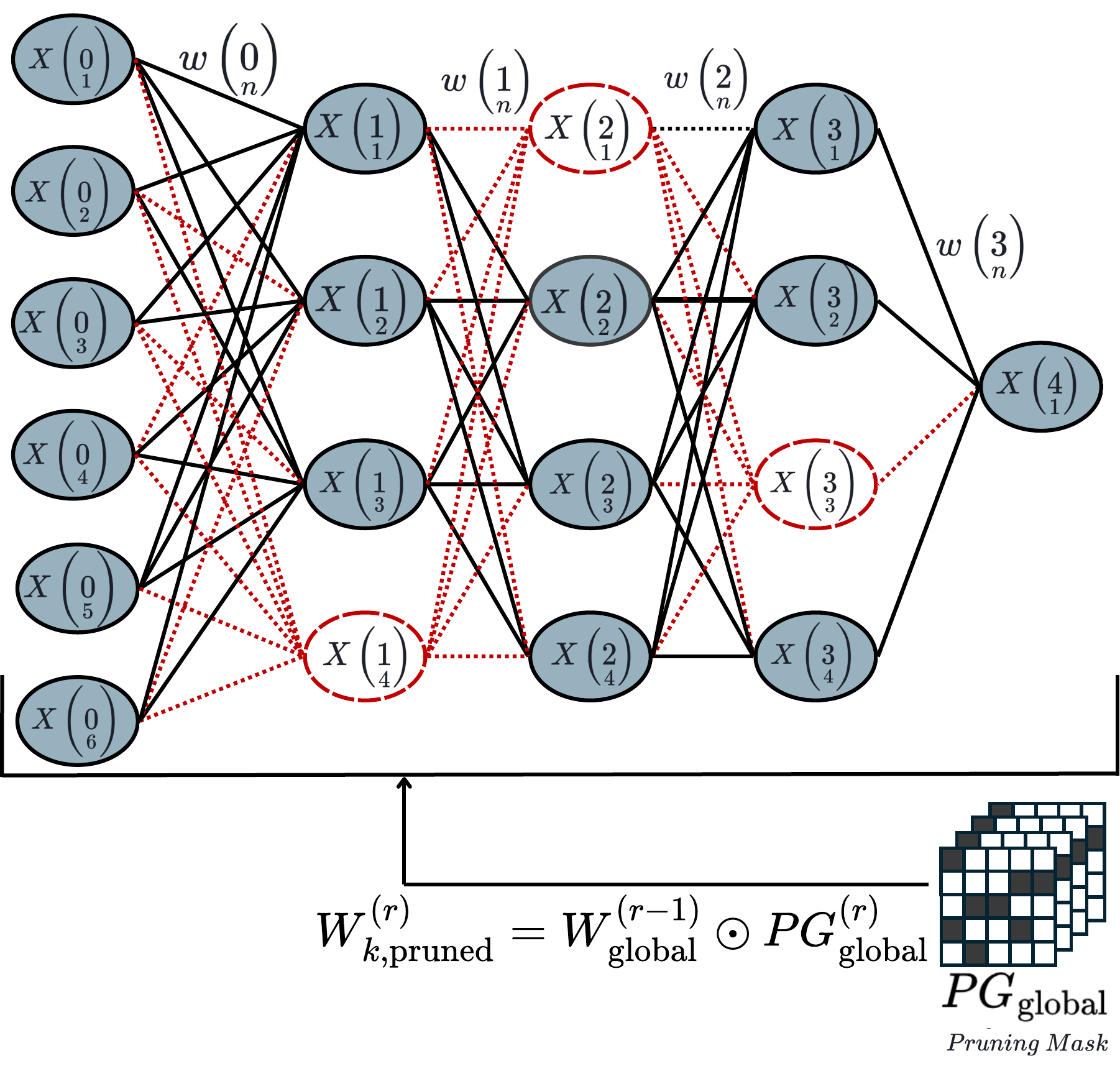}
    \caption{The \texttt{AutoFLIP} hybrid pruning mechanism. The global pruning mask simultaneously removes individual weights (unstructured, dotted lines) and, consequently, entire neurons when all their connections are removed (structured, highlighted nodes).}
    \label{fig:NNp}
\end{figure}

For a convolutional layer $l$, removing $\Delta F_l$ filters reduces the computational cost by:

\begin{equation}
    \Delta \text{FLOPs}_l = 2 \times \Delta F_l \times K_l^2 \times C_{l-1} \times H_l \times W_l.
\end{equation}
This FLOPs reduction enhances execution speed and energy efficiency, making complex models viable on resource-constrained devices. Figure \ref{fig:NNp} illustrates this hybrid approach.

\subsection{The Proposed \texttt{AutoFLIP} Framework}
\label{subsec:autoflip_framework}

The framework's operation, as detailed in Algorithm~\ref{alg:AutoFLIP}, proceeds as follows.

\begin{algorithm}[h]
    \caption{\texttt{AutoFLIP} Algorithm}
    \label{alg:AutoFLIP}
    \begin{algorithmic}[1]
    
    \State \textbf{Initialize:} Global model \( W_\text{global}^{(0)} \), pruning threshold \( T_p \).
    
    \Statex
    \begin{tcolorbox}[colback=phaseOneBlue!15, colframe=phaseOneBlue, sharp corners, boxrule=0.5pt, title=\textbf{Phase 1: Federated Loss Exploration}]
        \State Server selects exploration clients $C_\text{exp}$ and distributes $W_\text{global}^{(0)}$.
        \For{each client $i \in C_\text{exp}$ in parallel}
            \State Compute local deviation scores $\{\delta_{i,m}^2\}$  (Eq. \ref{eq:deviation}).
            \State Send scores to Server.
        \EndFor
        \State Server computes initial guidance matrix $G_{\text{global}}^{(0)}$ (Eq. \ref{eq:norm_deviation}).
    \end{tcolorbox}
    
    \Statex
    \begin{tcolorbox}[colback=phaseTwoRed!15, colframe=phaseTwoRed, sharp corners, boxrule=0.5pt, title=\textbf{Phase 2: Dynamic Adaptation during Training}]
        \For{each round $r = 1$ to $R$}
            \State Server selects a set of $K$ clients, $\mathcal{S}_r$.
            
            \Statex \textit{Server-Side Mask Generation:}
            \State Generate pruning mask $PG_{\text{global}}^{(r)}$ from $G_{\text{global}}^{(r-1)}$ using Eq. \ref{eq:binarization}.
            
            \For{each client $i \in \mathcal{S}_r$ in parallel}
                \State Receive $W_\text{global}^{(r-1)}$ and $PG_{\text{global}}^{(r)}$.
                \State $W_{i, \text{pruned}}^{(r-1)} \leftarrow W_\text{global}^{(r-1)} \odot PG_{\text{global}}^{(r)}$.
                \State $W_{i, \text{updated}} \leftarrow \text{LocalUpdate}(W_{i, \text{pruned}}^{(r-1)}, D_i)$.
                \State $\Delta W_i^{(r)} \leftarrow W_{i, \text{updated}} - W_{i, \text{pruned}}^{(r-1)}$.
                \State Send sparse update $\Delta W_i^{(r)} \odot PG_{\text{global}}^{(r)}$ to Server.
            \EndFor
            
            \Statex \textit{Server-Side Aggregation and Guidance Update:}
            \State Aggregate model: $W_\text{global}^{(r)} \leftarrow W_\text{global}^{(r-1)} + \frac{1}{K} \sum_{i \in \mathcal{S}_r} (\Delta W_i^{(r)} \odot PG_{\text{global}}^{(r)})$.
            \State Compute agreement score $A^{(r)}$ and importance $I^{(r)}$ from $\{\Delta W_i^{(r)}\}_{i \in \mathcal{S}_r}$ using Eqs. \ref{eq:agreement} and \ref{eq:importance}.
            \State Update guidance for next round via adaptive EMA using Eq. \ref{eq:adaptive_ema_update}.
        \EndFor
    \end{tcolorbox}
    
    \end{algorithmic}
\end{algorithm}

\noindent\textbf{Initialization (Line 1):} The server initializes the global model $W_\text{global}^{(0)}$ and the user-defined pruning threshold $T_p$.

\noindent\textcolor{phaseOneBlue}{\textbf{Phase 1: Federated Loss Exploration (Lines 2-6):}} This phase is executed once. A set of clients $C_\text{exp}$ is selected to perform local training and report their parameter loss deviations. The server aggregates these reports to compute the initial guidance matrix $G_{\text{global}}^{(0)}$, which provides a foundational estimate of parameter importance.

\noindent\textcolor{phaseTwoRed}{\textbf{Phase 2: Dynamic Adaptation (Lines 8-21):}} The main training process iterates for $R$ communication rounds.
\begin{itemize}
    \item At the start of each round $r$, the server generates the binary pruning mask $PG_{\text{global}}^{(r)}$ by thresholding the guidance matrix from the previous round, $G_{\text{global}}^{(r-1)}$ (Line 11).
    \item The server distributes the global model and the new mask to the selected clients. Each client applies the mask to create a sparse local model before training on its local data (Lines 13-15).
    \item After local training, each client calculates the update relative to the pruned model and transmits only the sparse update back to the server, reducing communication costs (Lines 16-17).
    \item The server aggregates the sparse updates to produce the new global model, $W_\text{global}^{(r)}$ (Line 19). It then computes the round-specific importance scores $I^{(r)}$ and agreement scores $A^{(r)}$ based on the received updates. Finally, it uses these scores to update the guidance matrix to $G_{\text{global}}^{(r)}$ via the adaptive EMA rule, which will inform pruning in the subsequent round (Lines 20-21).
\end{itemize}

\begin{figure}[!h]
    \centering
    \includegraphics[width=\columnwidth]{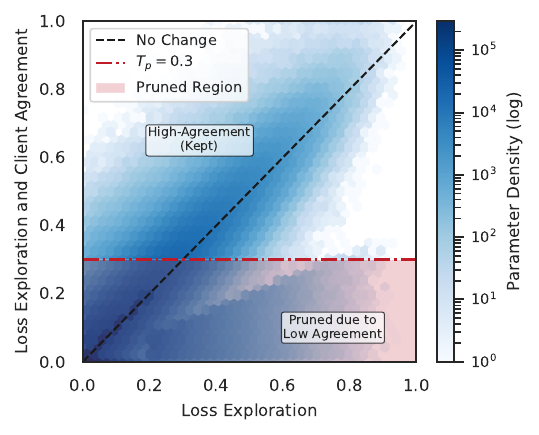}
    \caption{\rebuttal{Synthesis of the \texttt{AutoFLIP} pruning process for the Six-layer CNN on MNIST. The plot shows the relationship between a parameter's initial importance (x-axis) and its final importance after adaptation via client agreement (y-axis). The $T_p$ threshold acts as the final decision boundary, pruning all parameters in the shaded region. This visualizes the two primary reasons for pruning: low initial importance or low client agreement during training.}}
    \label{fig:guidance_synthesis}
\end{figure}

\rebuttal{This cycle ensures the pruning strategy is continuously informed by client consensus. Figure~\ref{fig:guidance_synthesis} provides a holistic view of this mechanism. It simulates the importance evolution for the $\approx7.6$ million parameters in the Six-layer CNN on the MNIST dataset. The plot maps each parameter's initial importance from loss exploration (x-axis) against its final, agreement-adapted importance (y-axis). This reveals two key dynamics. The dense diagonal represents stable, high-agreement parameters that are correctly retained. In contrast, an off-diagonal cluster in the lower right highlights our adaptation phase. Here, parameters that initially appeared important are demoted due to low client agreement. The pruning threshold, $T_p$, acts as the final decision boundary. It removes all parameters in the shaded region. This ensures the final subnetwork is composed only of parameters that are both salient and consistently agreed upon by clients.}

\section{Experiments}
\label{subsec:exp}
\subsection{Experimental Setup}
Inspired by \cite{Hahn2022}, we benchmark \texttt{AutoFLIP} across a diverse suite of six federated learning tasks to evaluate its robustness and generality. We explore multiple data partitioning approaches to create strongly non-IID conditions:
\textbf{\textit{Pathological non-IID:}} This scenario involves clients with data from only two distinct classes. We employ the MNIST dataset with a six-layer CNN and CIFAR-10 with EfficientNet-B3.

\textbf{\textit{Dirichlet-based non-IID:}} Here, we use a Dirichlet distribution to distribute data among clients, resulting in varying class counts per client to simulate realistic label skew. This approach is applied to the fine-grained CIFAR100 dataset using a ResNet model and the more complex Tiny ImageNet dataset with MobileNetV2.

\textbf{\textit{LEAF non-IID:}} Adopting the LEAF benchmark \cite{caldas2019leaf}, we evaluate \texttt{AutoFLIP} on the FEMNIST and Shakespeare datasets. For FEMNIST, a CNN architecture is used, while a two-layer LSTM model is employed for Shakespeare.

We evaluate \texttt{AutoFLIP} against \texttt{FedAvg} (no compression) and state-of-the-art algorithms such as \texttt{PruneFL}~\cite{jiang_model_2023}, \texttt{EFLPrune}~\cite{Wu2023}, and \texttt{FedMask}~\cite{fedmask}. The general experimental setup is summarized in Table~\ref{tab:hyperparameters}.

\begin{table}[h]
\caption{Experimental Setup Hyperparameters}
\label{tab:hyperparameters}
\centering
\small 
\renewcommand{\arraystretch}{0.9} 
\setlength{\tabcolsep}{4pt}      
\begin{tabular}{l p{0.55\columnwidth}} 
\toprule
\textbf{Parameter} & \textbf{Value} \\
\midrule
Clients (\(C\)) & 20 (see Sec. V-A for LEAF datasets) \\
Clients per round (\(K\)) & 5 (LEAF: 20) \\
Total Rounds (\(R\)) & 200 \\
Batch Size (\(B\)) & 32 \\
Local Epochs (\(E\)) & 10 \\
Learning Rate (\(\eta\)) & 0.001 \\
\midrule
 Adam Optimizer &  weight decay, server momentum =0.9 \\
\midrule
Exploration & \(E_\text{exp}\)=150 (patience=20), \(T_p\)=0.3 \\
\bottomrule
\end{tabular}
\end{table}

Data is divided into 80\% for training and 20\% for testing. Global model performance is assessed by the average prediction accuracy on the test sets. To ensure statistical validity, each experiment is repeated 10 times. We measure the compression rate to evaluate model size reduction and its impact. The computational effort, measured in FLOPs, is estimated using the \texttt{THOP: PyTorch-OpCounter} function~\cite{zhu2022thop} in \texttt{PyTorch}. Experiments were conducted on a machine equipped with an Intel Xeon X5680 CPU, 128 GB DDR4 RAM, and an NVIDIA TITAN X GPU.

\subsection{Global Accuracy Convergence Results}

\begin{figure*}[!t]
    \centering
    \includegraphics[width=\textwidth]{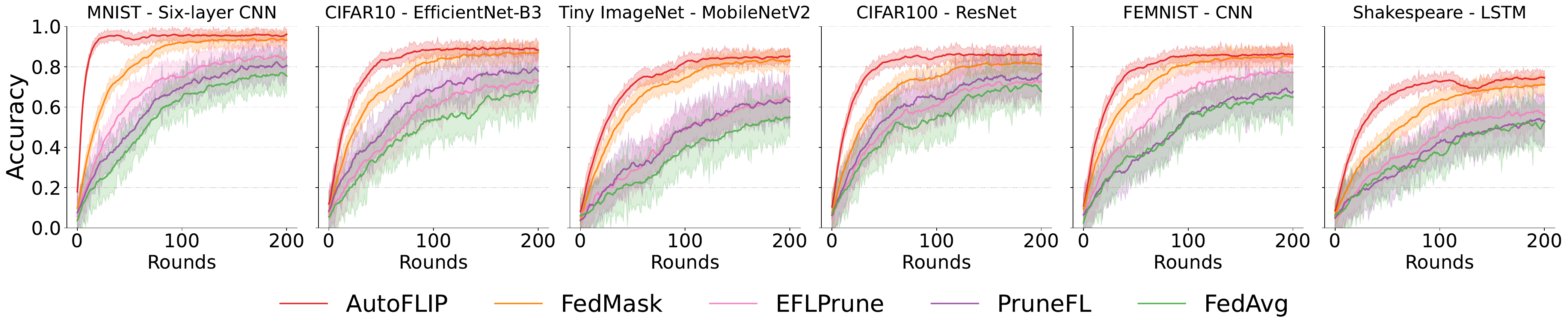}
    \caption{Global accuracy convergence of \texttt{AutoFLIP} against FedAvg~\cite{mcmahan2023communicationefficient}, PruneFL~\cite{jiang_model_2023}, EFLPrune \cite{Wu2023} and \texttt{FedMask} \cite{fedmask}. }
    \label{fig:accuracy}
\end{figure*}

\begin{table*}[!t]
    \caption{Final global model accuracy comparison across all datasets and models. Results are presented as mean \( \pm \) standard deviation. Best results are highlighted.}
    \label{tab:final_accuracy_eflprune}
    \centering
    \resizebox{\textwidth}{!}{%
    \begin{tabular}{lcccccc}
        \toprule
        \textbf{Algorithm} & \textbf{MNIST - CNN} & \textbf{CIFAR10 - EffNetB3} & \textbf{Tiny ImageNet - MobileNetV2} & \textbf{CIFAR100 - ResNet} & \textbf{FEMNIST - CNN} & \textbf{Shakespeare - LSTM} \\
        \midrule
        \textbf{\texttt{FedAvg}} & $0.803 \pm 0.051$ & $0.751 \pm 0.062$ & $0.687 \pm 0.065$ & $0.724 \pm 0.061$ & $0.682 \pm 0.064$ & $0.559 \pm 0.062$ \\
        \textbf{\texttt{PruneFL}} & $0.835 \pm 0.046$ & $0.796 \pm 0.056$ & $0.708 \pm 0.058$ & $0.763 \pm 0.057$ & $0.722 \pm 0.058$ & $0.581 \pm 0.056$ \\
        \textbf{\texttt{EFLPrune}} & $0.857 \pm 0.042$ & $0.774 \pm 0.053$ & $0.749 \pm 0.051$ & $0.746 \pm 0.053$ & $0.795 \pm 0.051$ & $0.613 \pm 0.052$ \\
        \textbf{\texttt{FedMask}} & $0.936 \pm 0.021$ & $0.878 \pm 0.026$ & $0.842 \pm 0.028$ & $0.821 \pm 0.027$ & $0.854 \pm 0.026$ & $0.727 \pm 0.028$ \\
        \midrule
        \textbf{\texttt{AutoFLIP}} & \best{$0.955 \pm 0.018$} & \best{$0.893 \pm 0.022$} & \best{$0.855 \pm 0.024$} & \best{$0.869 \pm 0.023$} & \best{$0.862 \pm 0.021$} & \best{$0.748 \pm 0.025$} \\
        \bottomrule
    \end{tabular}%
    }
\end{table*}

The complete learning curves for all methods across the six benchmarks are presented in Figure~\ref{fig:accuracy}, with final accuracy metrics summarized in Table~\ref{tab:final_accuracy_eflprune}. The results show that \texttt{AutoFLIP} consistently achieves a superior or competitive trade-off between convergence speed and final accuracy, though the margin of improvement varies depending on the task's complexity and data modality.

On the MNIST and CIFAR100 datasets, \texttt{AutoFLIP} establishes a performance advantage. As shown in Table~\ref{tab:final_accuracy_eflprune}, it achieves a final accuracy of 0.955 on MNIST, an improvement over the next-best baseline, \texttt{FedMask} (0.936). 

However, the performance landscape is more nuanced on more challenging benchmarks. On Tiny ImageNet, the advantage narrows, with \texttt{AutoFLIP} (0.855) achieving only a marginal gain over \texttt{FedMask} (0.842). Most notably, on the highly non-IID Shakespeare dataset, \texttt{AutoFLIP} (0.748) performs nearly on par with \texttt{FedMask} (0.727) and exhibits a distinct late-stage performance dip, highlighting the challenges of sequential, non-IID text data.

The baseline methods exhibit diverse and illustrative behaviors. As depicted in Figure~\ref{fig:accuracy}, the lower-performing algorithms often suffer from pronounced initial instability. For instance, in the CIFAR10 experiment, \texttt{PruneFL} experiences a severe performance drop in the early rounds from which it struggles to recover. In contrast, on the same dataset, \texttt{FedAvg} suffers from a late-stage dip, failing to maintain its peak accuracy. These varied pathologies underscore the complexity of the federated optimization landscape. The consistently high variance and lower final accuracy of the baseline methods further validate the stability and effectiveness of our proposed approach.

\subsection{Pruning Efficiency Results}

Beyond improving final model accuracy, a primary objective of \texttt{AutoFLIP} is to enhance the overall efficiency of the FL process. We evaluate this by analyzing the three-way trade-off between final model accuracy, total communication cost, and total computational cost (FLOPs). 

\begin{figure}[!t]
    \centering
    \includegraphics[width=\columnwidth]{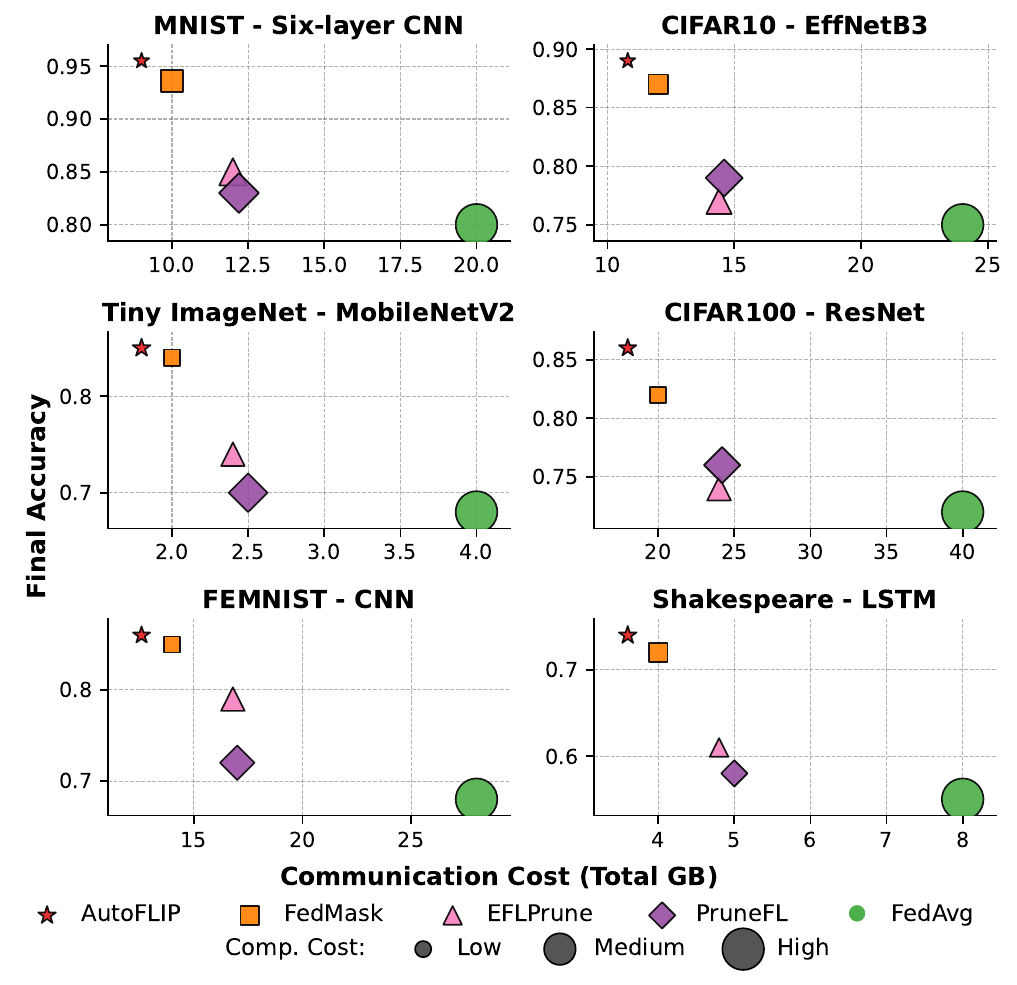}
    \caption{Accuracy vs. Efficiency Trade-off. The plot visualizes final model accuracy against total communication cost. The size of each marker represents the total computational cost (FLOPs), with smaller bubbles indicating higher efficiency. \texttt{AutoFLIP} consistently demonstrates a superior trade-off by occupying the top-left region with a smaller marker size.}
    \label{fig:tradeoff}
\end{figure}

\rebuttalsecond{Figure~\ref{fig:tradeoff} provides a comprehensive visualization of this trade-off. To ensure a fair comparison, the total communication and computational costs reported are inclusive of the resources consumed during the initial one-time federated exploration phase.}
In these plots, the ideal performance is located in the top-left corner, representing high accuracy for low communication cost. The size of each marker is inversely proportional to the computational cost; a smaller bubble indicates a more computationally efficient method. Across all six diverse benchmarks, \texttt{AutoFLIP} consistently occupies the most favorable position, demonstrating its ability to achieve a superior balance of all three objectives.

\subsubsection{Computational Effort}
The emergent structured pruning in \texttt{AutoFLIP} is key to reducing the computational burden on client devices.\rebuttal{While Figure~\ref{fig:tradeoff} visualizes this computational saving as a relative marker size to illustrate the three-way trade-off among the SoTA baselines, Table~\ref{tab:summary_flops} provides the quantitative breakdown of these savings compared to the classic \textit{FedAvg}.} As detailed in Table \ref{tab:summary_flops}, our approach yields a substantial reduction in FLOPs across all tested architectures. We observe a greater than 40\% reduction for most models, with a peak reduction of 56.5\% for the FEMNIST-CNN. These results confirm that \texttt{AutoFLIP} produces models that are not only smaller but also significantly faster and more energy-efficient to train and use for inference on-device.

\begin{table}[!t]
    \caption{This table compares the original and reduced FLOPs after applying \texttt{AutoFLIP} with compression rate and the \% reduction shows the efficiency achieved.}
    \label{tab:summary_flops}
    \centering
    \begin{tabularx}{\columnwidth}{lXXXX}
        \toprule
        \textbf{Model} & \textbf{Compr. Rate} & \textbf{Orig. (GFLOPS)} & \textbf{Red. (GFLOPS)} & \textbf{\% Red.} \\
        \midrule
        Six-layer CNN & 1.74x & 13.3 & 5.4 & 59.4\% \(\downarrow\) \\
        EfficientNet-B3 & 2.1x & 15.7 & 7.2 & 54.1\% \(\downarrow\) \\
        MobileNetV2 & 1.9x & 0.3 & 0.16 & 46.7\% \(\downarrow\) \\
        ResNet & 1.58x & 7.8 & 4.1 & 47.4\% \(\downarrow\) \\
        FEMNIST-CNN & 1.8x & 19.4 & 10.1 & 48.0\% \(\downarrow\) \\
        LSTM & 1.8x & 10.1 & 4.4 & 56.4\% \(\downarrow\) \\
        \bottomrule
    \end{tabularx}
\end{table}

\subsubsection{Communication Costs}
\begin{table}[!t]
    \caption{Comparison of communication costs with and without \texttt{AutoFLIP}. The metric is rounds to a target accuracy of 75\%. N/A indicates the method failed to reach the target within 200 rounds.}
    \label{tab:communication_costs}
    \centering
    \resizebox{\columnwidth}{!}{%
    \begin{tabular}{lrrccc}
        \toprule
        \multirow{2}{*}{\textbf{Model}} & \multicolumn{2}{c}{\textbf{Rounds to Target (75\%)}} & \multicolumn{2}{c}{\textbf{Est. Cost (GB)}} & \multirow{2}{*}{\textbf{\% Reduction}} \\
        \cmidrule(lr){2-3} \cmidrule(lr){4-5}
        & \textbf{AutoFLIP} & \textbf{Baseline} & \textbf{AutoFLIP} & \textbf{Baseline} & \\
        \midrule
        Six-layer CNN & 20 & 90 & 10.0 & 45.0 & 77.8\% \(\downarrow\) \\
        EfficientNet-B3 & 45 & 190 & 27.0 & 114.0 & 76.3\% \(\downarrow\) \\
        MobileNetV2 & 80 & N/A & 8.0 & N/A & - \\
        CIFAR100 - ResNet & 60 & 180 & 60.0 & 180.0 & 66.7\% \(\downarrow\) \\
        FEMNIST-CNN & 65 & 150 & 45.5 & 105.0 & 56.7\% \(\downarrow\) \\
        Shakespeare & N/A & N/A & N/A & N/A & - \\
        \bottomrule
    \end{tabular}
    }
\end{table}
To provide a practical measure of communication efficiency, we evaluate the number of rounds required for each method to achieve a challenging target accuracy of 75\%. This time-to-accuracy metric is crucial for deployments where achieving a high-performance model quickly is a primary goal. The results, presented in Table \ref{tab:communication_costs}, demonstrate that \texttt{AutoFLIP} substantially reduces the communication rounds required to reach this performance threshold. For instance, on the CIFAR10 dataset, \texttt{AutoFLIP} reached the 75\% target in only 45 rounds, while the baseline \texttt{FedAvg} required 190 rounds. This represents a 4.2x acceleration and a corresponding 76.3\% reduction in data transferred.

\rebuttalthird{
\subsubsection{Complexity Analysis}
To formalize the efficiency gains, Table~\ref{tab:complexity} contrasts the theoretical complexity of \texttt{AutoFLIP} against the baselines. We distinguish between theoretical FLOPs (assuming sparse hardware) and practical execution on standard dense hardware.This comparison clarifies a critical distinction. Baselines like \texttt{FedMask} and \texttt{PruneFL} rely on unstructured masking, where weight matrices retain original dimensions ($M \times M$), often yielding no FLOPs reduction on standard dense hardware ($O(M \cdot E)$). In contrast, \texttt{AutoFLIP} employs emergent structured pruning to reduce dimensions to $M_{struct}$, guaranteeing accelerated execution ($O(M_{struct} \cdot E)$) on any platform. Additionally, unlike the iterative or memory-intensive overheads of baselines, \texttt{AutoFLIP}'s single exploration cost at $t=0$ is effectively amortized over the full training duration.}

\rebuttal{
\begin{table}[h]
\caption{Complexity and Overhead Comparison.}
\label{tab:complexity}
\centering
\resizebox{\columnwidth}{!}{%
\begin{tabular}{l|ccc}
\toprule
\textbf{Method} & \textbf{Comm. ($O$)} & \textbf{Comp. ($O$)} & \textbf{Specific Overhead} \\
\midrule
\texttt{FedAvg} & $M$ & $M \cdot E$ & None \\
\texttt{PruneFL} & $M(1-S)$ & $M \cdot E \cdot (1-S)^*$ & Initial Pre-Pruning Phase \\
\texttt{FedMask} & $M_{mask}$ & $M \cdot E$ & Frozen Weights + Mask Learning \\
\textbf{\texttt{AutoFLIP}} & $\mathbf{M(1-S)}$ & $\mathbf{M_{struct} \cdot E}$ & One-time Exploration ($t=0$) \\
\bottomrule
\end{tabular}%
}
\vspace{1mm}
\scriptsize{\textit{*Unstructured pruning requires specialized sparse hardware/libraries to realize computational speedups. \texttt{AutoFLIP} ($M_{struct}$) reduces physical matrix dimensions, accelerating execution on standard hardware.}}
\end{table}
}

\begin{takeawaybox}{Takeaway 2: Key Results}
\texttt{AutoFLIP} delivers a clear win-win: it drastically cuts on-device computation and network communication by over 50\% while simultaneously achieving higher accuracy than current state-of-the-art methods, especially in challenging, real-world (non-IID) scenarios.
\end{takeawaybox}

\subsection{Robustness to Noisy Clients}
\label{sec:noyse}

\rebuttal{ Real-world clients can produce noisy updates due to data corruption or transmission errors. We hypothesized our Client Agreement Score (Eq.~\ref{eq:agreement}) provides inherent robustness by filtering these non-consensus updates. To validate this, we extended the MNIST and CIFAR10 experiments using the pathological non-IID baseline from Section V.A. We then conducted two new experiments: \textbf{(1)} sensitivity to update-level Gaussian noise and \textbf{(2)} robustness to data-level label noise.}

\begin{figure}[!t]
    \centering
    \includegraphics[width=\columnwidth]{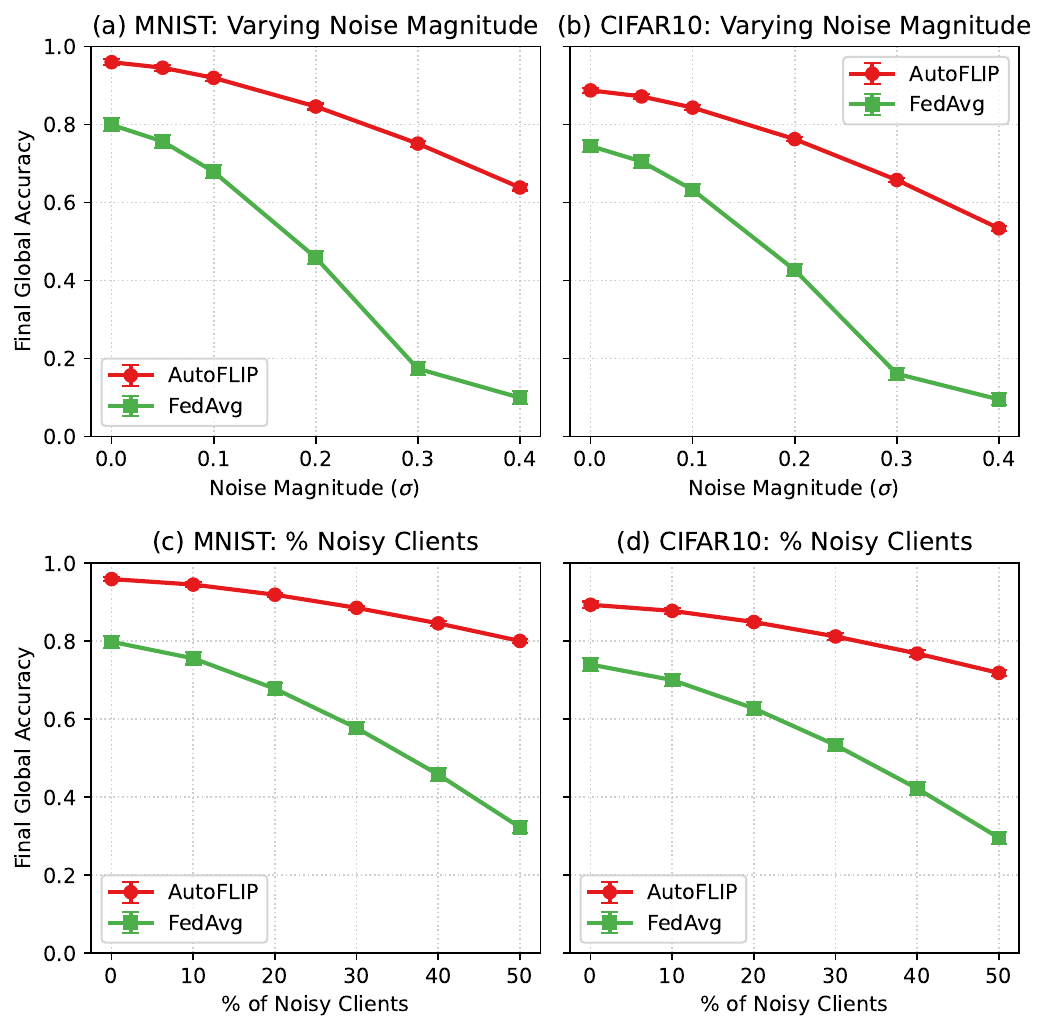}
    \caption{\rebuttal{Robustness Sensitivity Analysis. We sweep (a, b) the noise magnitude ($\sigma$) with a 20\% noisy client scope, and (c, d) the client scope (\%) with a 0.1 $\sigma$. The results show \texttt{AutoFLIP}'s final accuracy (red) degrades gracefully, while \texttt{FedAvg}'s (green) collapses.}}
    \label{fig:noise_sensitivity}
\end{figure}

\rebuttal{\textbf{(1)} We analyzed sensitivity to update-level noise by injecting Gaussian noise ($\mathcal{N}(0, \sigma^2)$) into client updates. We tested various noise magnitudes ($\sigma$) and percentages of noisy clients. The results in Figure~\ref{fig:noise_sensitivity} show \texttt{FedAvg}'s (green line) accuracy collapses non-linearly. For example, \texttt{FedAvg}'s accuracy drops to near chance-level ($<0.3$) with only $40\%$ noisy clients (Fig.~\ref{fig:noise_sensitivity}c, d). In contrast, \texttt{AutoFLIP}'s (red line) accuracy degrades far more gracefully. It maintained $>0.8$ accuracy on MNIST and $>0.75$ on CIFAR10 in the same scenario.}

\begin{figure}[!t]
    \centering
    \includegraphics[width=\columnwidth]{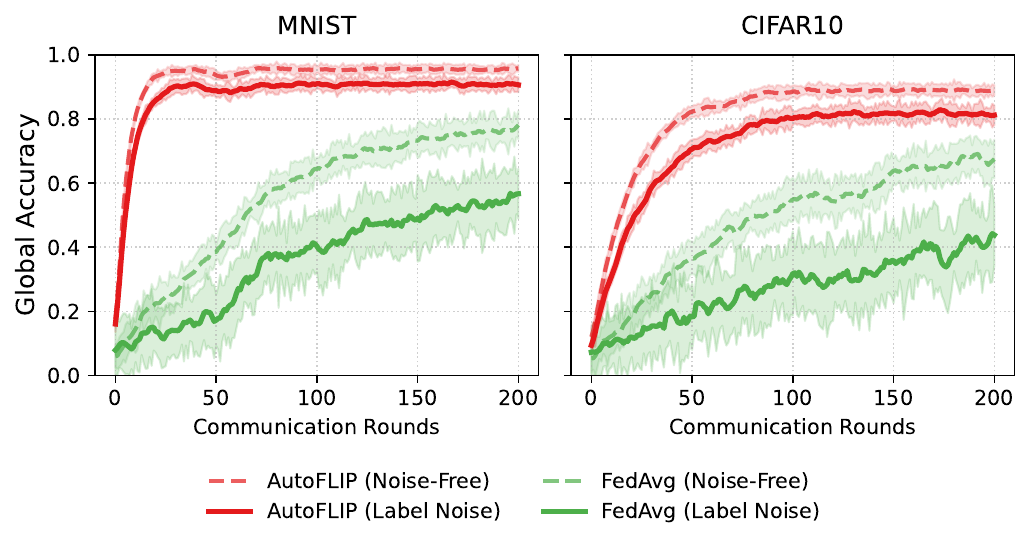}
    
    \caption{\rebuttal{Robustness to Data-Level (Label) Noise. On 20\% of clients, 30\% of local data labels were flipped. On MNIST and on CIFAR10, \texttt{FedAvg} (green) suffers severe accuracy degradation from the divergent updates. \texttt{AutoFLIP} (red) identifies these as low-agreement updates and filters them, maintaining high stability.}}
    \label{fig:label_noise}
\end{figure}

\rebuttal{\textbf{(2)} Second, we tested robustness to a \textit{data-level} threat. We simulated $20\%$ of clients training on data with $30\%$ flipped labels. This creates confidently wrong clients with internally consistent but globally divergent updates. The results in Figure~\figref{fig:label_noise} are clear. \texttt{FedAvg} (green line) is severely impacted. Its convergence is unstable (wide variance shadow) and its final MNIST accuracy collapses to $\approx0.6$. \texttt{AutoFLIP} (red line), however, is highly resilient. Its Client Agreement Score (Eq.~\ref{eq:agreement}) identifies these divergent updates as outliers. Its adaptive EMA (Eq.~\ref{eq:adaptive_ema_update}) then filters their influence. This results in stable convergence (tight variance shadow) to a high final accuracy (e.g., $\approx0.9$ on MNIST), proving it can handle confidently wrong clients.}

\section{Ablation Study}
\label{sec:ablation}
\subsection{Impact of Exploration Parameters \( C_\text{exp} \) and \( E_\text{exp} \)}
\label{subsec:abl1}

\begin{figure*}[!t]
    \centering
    \includegraphics[width=0.9\textwidth]{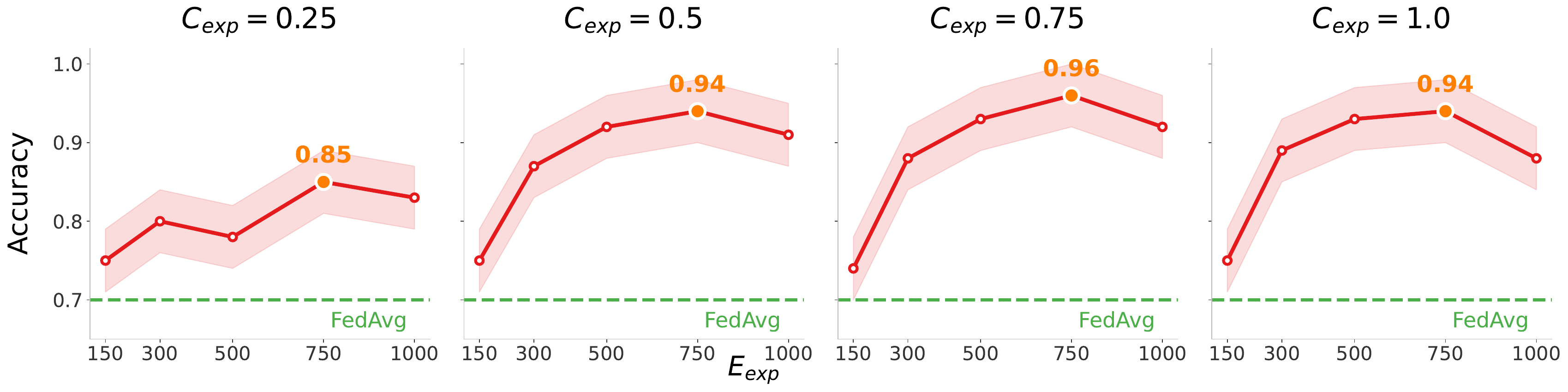}
    \caption{Ablation study on the number of exploration clients \( C_\text{exp} \) and the number of exploration epochs \( E_\text{exp} \) for FEMNIST under the LEAF non-IID scenario, based accuracy. The x-axis represents \( E_\text{exp} \), and each plot corresponds to a different \( C_\text{exp} \).}
    \label{fig:ablabl}
\end{figure*}

We conduct an ablation study to assess the sensitivity of \texttt{AutoFLIP} to the parameters \( C_\text{exp} \) and \( E_\text{exp} \). The number of explorer clients, \( C_\text{exp} \), influences the comprehensiveness of the global guidance matrix in capturing the intricacies of the clients' loss landscapes. The depth of the exploration phase, quantified by the number of exploration epochs, \( E_\text{exp} \), affects the understanding of the loss function surface.
We examine how the average accuracy and loss for the global model vary for \( C_\text{exp} \in \{0.25, 0.5, 0.75, 1.0\} \) and \( E_\text{exp} \in \{150, 300, 500, 750, 1000\} \). This evaluation is conducted on the FEMNIST dataset under the LEAF non-IID scenario, as depicted in Figure~\ref{fig:ablabl}.

Our findings reveal that increasing the number of exploration clients \( C_\text{exp} \) enhances the robustness of the global guidance matrix, resulting in improved initial accuracy. However, even a conservative value of \( C_\text{exp} \) can boost accuracy, indicating that \texttt{AutoFLIP} benefits from a diverse set of explorer clients. Additionally, increasing the number of exploration epochs \( E_\text{exp} \) generally leads to improvements in accuracy. Beyond a certain point (e.g., \( E_\text{exp} = 300 \)), the gains diminish, indicating a saturation point where additional exploration yields minimal benefits.\rebuttalthird{These results suggest scalability for large IoT networks. If the number of explorers is a fixed fraction $p$ of the nodes,
$C_{\mathrm{exp}} = pC$, then the complexity of the exploration phase is
$\mathcal{O}(C_{\mathrm{exp}}) = \mathcal{O}(C)$, which grows linearly with the network size.}

\subsection{Impact of Pruning Threshold \( T_\text{p} \)}
\label{subsec:abl2}

\begin{figure*}[!t]
    \centering
    \includegraphics[width=0.9\textwidth, height=0.35\textwidth]{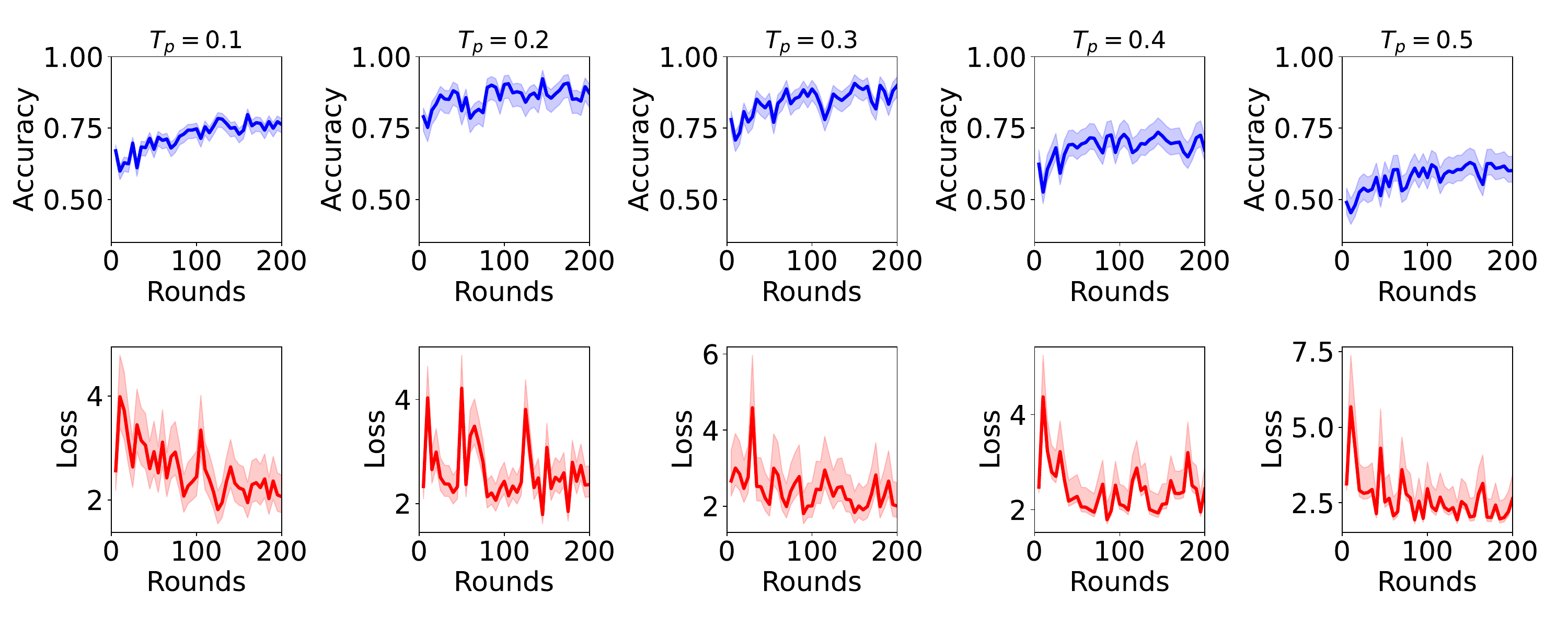}
    \caption{Ablation study on the pruning threshold \( T_\text{p} \) for CIFAR10 under a pathological non-IID data distribution, based on average accuracy (top) and loss (bottom).}
    \label{fig:ablation_tp}
\end{figure*}

We evaluate the sensitivity of \texttt{AutoFLIP} to the pruning threshold parameter, \( T_\text{p} \). Specifically, we assess how the average accuracy and loss vary for \( T_\text{p} \in \{0.1, 0.2, 0.3, 0.4, 0.5\} \) on the CIFAR10 dataset under the Pathological non-IID scenario, as shown in Figure~\ref{fig:ablation_tp}.

Our results indicate that setting \( T_p = 0.3 \) strikes the best balance between compression rate and model performance, achieving high accuracy while maintaining a significant reduction in model size. Lowering \( T_p \) retains more parameters, leading to better accuracy but with reduced compression benefits. Conversely, increasing \( T_p \) results in more aggressive pruning, which can degrade model accuracy. This analysis highlights how the \(T_p\) parameter serves as a direct and effective control knob, allowing an IBN system to enforce a specific policy that balances resource efficiency with model performance.

\section{Conclusion and Future Work}
\label{sec:conclusion}

In this work, we introduced \texttt{AutoFLIP}, a federated learning framework designed to bridge the gap between high-level network intents and the practical deployment of AI models in heterogeneous IoT environments. By leveraging a dynamic guidance system that evolves based on client loss exploration and update agreement to inform an adaptive hybrid pruning strategy, \texttt{AutoFLIP} successfully translates an intent for efficiency into a concrete, optimized, and hardware-aware model. Our extensive experiments demonstrated that this approach is both practical and highly effective. We showed that \texttt{AutoFLIP} consistently discovers superior sparse subnetworks, leading to significant reductions in computational overhead (averaging 52\%) and accelerating convergence to target accuracies. Critically, this efficiency was achieved while improving model accuracy over state-of-the-art methods in challenging non-IID settings (e.g., +4.9\% on CIFAR100). Even on the most difficult benchmarks, such as Shakespeare, where its accuracy advantage was minimal, it demonstrated greater stability and reliability than baseline methods. \rebuttal{Ultimately, \texttt{AutoFLIP} provides a grounded solution that actualizes IBN for distributed machine learning. By successfully translating a high-level, declarative intent (the $T_p$ threshold) into an optimized, self-adapting, and hardware-aware network configuration, it provides a practical method for facilitating more scalable and robust AI applications in resource-constrained networks.}

\noindent \textit{Future Research Directions}: 
While our results are demonstrated in a single-server setting, the principle of collaborative exploration opens several exciting avenues for future research. \rebuttalsecond{Extending this framework to hierarchical multi-server environments is a clear next step, where intermediate servers could learn ``regional'' guidance maps, allowing a top-level orchestrator to cluster them and assign different, fine-grained $T_p$ intents based on the network's discovered topology.} \rebuttalsecond{Furthermore, investigating the privacy implications of the exploration phase (e.g., from sharing $\delta_{i,m}^2$ scores) and developing formal safeguards (e.g., differential privacy~\cite{wei2019federatedlearningdifferentialprivacy} or multi-party computation~\cite{yuca2026ppflex}) will be crucial for real-world deployment.} \rebuttalthird{We hypothesize that sharing squared deviations $\delta_{i,m}^2$ presents a lower leakage risk than raw gradients due to temporal aggregation (masking single-batch contributions) and directional obfuscation (removing sign information). Because our strategy relies on the relative importance ranking rather than absolute magnitudes, we can apply strict client-side clipping bounds to the $\delta_{i,m}^2$ scores to mitigate the risk of outlier leakage while preserving the utility of the generated pruning mask.} \rebuttal{The concept of using a federated network to ``map'' a problem's structure could be extended beyond pruning to other areas of automated model design, such as for large-scale foundation and diffusion models~\cite{bommasani2022opportunitiesrisksfoundationmodels, moleri2026federatedfactorygenerativeoneshotlearning}.}\\

\textit{Broader Impact}: \texttt{AutoFLIP} enhances sustainability and efficiency in FL by reducing the energy footprint of training DL models. However, deploying \texttt{AutoFLIP} requires careful consideration of ethical issues, including data privacy and biases. Proactive management and regulation are crucial to ensure its positive societal impact and responsible integration into critical fields.

\section*{Acknowledgements}
This research was partly funded by Honda Research Institute Europe. The authors would like to thank Hao Wang, Jacob de Nobel, Nikolay Matyunin, Nergiz Yuca, David Klindt, and Riccardo Cadei for their helpful suggestions during the preparation of this manuscript.

\bibliographystyle{IEEEtran}
\bibliography{bibliography}

\end{document}